\documentclass[11pt]{article}
\usepackage{konvens2018}
\usepackage{mathptmx}
\usepackage[scaled=.90]{helvet}
\usepackage{courier}
\usepackage{graphicx}
\usepackage{booktabs}
\usepackage{tabularx}
\usepackage{multirow}
\usepackage{todonotes}

\usepackage{url}
\usepackage{latexsym}

\title{Transfer Learning from LDA to BiLSTM-CNN for \\ Offensive Language Detection in Twitter}

\author{Gregor Wiedemann \qquad Eugen Ruppert \qquad Raghav Jindal \qquad Chris Biemann \\
Language Technology Group\\ 
Department of Informatics\\ 
University of Hamburg, Germany\\
 {\tt \{gwiedemann, ruppert, biemann\}@informatik.uni-hamburg.de} \\
 {\tt raghavjindal2003@gmail.com}}

\date{}

\begin{document}
\maketitle
\begin{abstract}
We investigate different strategies for automatic offensive language classification on German Twitter data. For this, we employ a sequentially combined BiLSTM-CNN neural network. Based on this model, three  transfer learning tasks to improve the classification performance with background knowledge are tested. We compare
1. Supervised category transfer: social media data annotated with near-offensive language categories,
2. Weakly-supervised category transfer: tweets annotated with emojis they contain,
3. Unsupervised category transfer: tweets annotated with topic clusters obtained by Latent Dirichlet Allocation (LDA).
Further, we investigate the effect of three different  strategies to mitigate negative effects of `catastrophic forgetting' during transfer learning.
Our results indicate that transfer learning in general improves offensive language detection.
Best results are achieved from pre-training our model on the unsupervised topic clustering of tweets in combination with thematic user cluster information.

\end{abstract}

\section{Introduction}

User-generated content in forums, blogs, and social media not only contributes to a deliberative exchange of opinions and ideas but is also contaminated with offensive language such as threats and discrimination against people, swear words or blunt insults. 
The automatic detection of such content can be a useful support for moderators of public platforms as well as for users who could receive warnings or would be enabled to filter unwanted content. 

Although this topic now has been studied for more than two decades, so far there has been little work on offensive language detection for German social media content. Regarding this, we present a new approach to detect offensive language as defined in the shared task of the GermEval 2018 workshop.\footnote{\url{https://projects.fzai.h-da.de/iggsa}} For our contribution to the shared task, we focus on the question how to apply transfer learning for neural network-based text classification systems.

In Germany, the growing interest in hate speech analysis and detection is closely related to recent political developments such as the increase of right-wing populism, and societal reactions to the ongoing influx of refugees seeking asylum \cite{Ross.2016}. Content analysis studies such as \newcite{InstituteforStrategicDialogue.2018} have shown that a majority of hate speech comments in German Facebook is authored by a rather small group of very active users (5\% of all accounts engaging in hate speech). The findings suggest that already such small groups are able to severely disturb social media debates for large audiences.

From the perspective of natural language processing, the task of automatic detection of offensive language in social media is complex due to three major reasons.
\textit{First}, we can expect `atypical' language data due to incorrect spellings, false grammar and non-standard language variations such as slang terms, intensifiers, or emojis/emoticons. For the automatic detection of offensive language, it is not quite clear whether these irregularities should be treated as `noise' or as a signal.
\textit{Second}, the task cannot be reduced to an analysis of word-level semantics only, e.g. spotting offensive keyterms in the data. Instead, the assessment of whether or not a post contains offensive language can be highly dependent on sentence and discourse level semantics, as well as subjective criteria. In a crowd-sourcing experiment on `hate speech' annotation, \newcite{Ross.2016} achieved only very low inter-rater agreement between annotators. Offensive language is probably somewhat easier to achieve agreement on, but still sentence-level semantics and context or `world knowledge' remains important. 
\textit{Third}, there is a lack of a common definition of the actual phenomenon to tackle. Published studies focus on `hostile messages', `flames', `hate speech', `discrimination', `abusive language', or `offensive language'. Although certainly overlapping, each of these categories has been operationalized in a slightly different manner. Since category definitions do not match properly, publicly available annotated datasets and language resources for one task cannot be used directly to train classifiers for any respective other task.

\paragraph{\textbf{Contribution:}} For the offensive language detection presented in this paper, our approach is to use semi-supervised text classification to address all of the three challenges. 
In order to account for atypical language, we use sub-word embeddings to represent word tokens, words unseen during training, misspelled words and ‘words’ specifically used in the context of social media such as emojis.
% \todo{requires test: fasttext vs. word2vec embeddings; should we compare to fasttext classification as a baseline?}
To represent complex sequence information from tweets, we use a neural network model combining recurrent (e.g. Long-Short term memory, LSTM) \cite{Hochreiter.1997} and convolutional (CNN) layers. Both learning architectures, LSTM and CNN, have already been employed successfully in similar text classification tasks such as sentiment analysis \cite{Kim.2014}.
% \todo{requires test: LSTM-CNN vs LSTM, CNN alone} 
We expect the combination of LSTM and CNN to be especially useful in the context of transfer learning.

The main contribution of this paper is to investigate potential performance contributions of transfer learning to offensive language detection. 
% \todo{requires test: different transfer sources.}
For this, we investigate three different approaches to make use of knowledge learned by one task to improve classification for our actual offensive language task. To pre-train our BiLSTM-CNN network, we employ
1. Supervised category transfer: social media data annotated with near-offensive language categories,
2. Weakly-supervised category transfer: tweets annotated with emojis they contain, and
3. Unsupervised category transfer: tweets annotated with topic clusters obtained by Latent Dirichlet Allocation (LDA) \cite{Blei.2003}.
% \todo{Test also LM transfer?}
Further, we investigate the effect of three different transfer learning strategies on the classification performance to mitigate the effect of `catastrophic forgetting'.\footnote{Catastrophic forgetting refers to the phenomenon that during supervised learning of the actual task in a transfer learning setup the update of model parameters can overwrite knowledge obtained by the previously conducted training task. This will eventually eliminate any positive effect of pre-training and knowledge transfer from background corpora.}
The results indicate that transfer learning on generic topic clusters of tweets derived from an LDA process of a large Twitter background corpus significantly improves offensive language detection. 

We present our findings in the following structure:  Section~\ref{sec:related} addresses related work to our approach. In Section~\ref{sec:data}, we introduce the details of the Germ\-Eval 2018 Shared Task together with our background corpora for knowledge transfer. In Section~\ref{sec:model}, we describe our BiLSTM-CNN model for text classification. Section~\ref{sec:transfer} introduces the different transfer learning setups we investigate. 
To evaluate these setups, we conduct a number of experiments for which results are presented in Section~\ref{sec:eval}. This section also contains a brief discussion of errors made by our model. 
Finally, we give some concluding remarks.

\section{Related Work}
\label{sec:related}

Automatic detection of offensive language is a well-studied phenomenon for the English language.
Initial works on the detection of `hostile messages' have been published already during the 1990s \cite{Spertus.1997}. An overview of recent approaches comparing the different task definitions, feature sets and classification methods is given by \newcite{Schmidt.2017}. 
A major step forward to support the task was the publication of a large publicly available, manually annotated dataset by Yahoo research \cite{Nobata.2016}. They provide a classification approach for detection of abusive language in Yahoo user comments using a variety of linguistic features in a linear classification model. One major result of their work was that learning text features from comments which are temporally close to the to-be-predicted data is more important than learning features from as much data as possible. This is especially important for real-life scenarios of classifying streams of comment data.
In addition to token-based features, \newcite{Xiang.2012} successfully employed topical features to detect offensive tweets. We will build upon this idea by employing topical data in our transfer learning setup.
Transfer learning recently has gained a lot of attention since it can be easily applied to neural network learning architectures. For instance, \newcite{Howard.2018} propose a generic transfer learning setup for text classification based on language modeling for pre-training neural models with large background corpora.
To improve offensive language detection for English social media texts, a transfer learning approach was recently introduced by \newcite{Felbo.2017}. Their `deepmoji' approach relies on the idea to pre-train a neural network model for an actual offensive language classification task by using emojis as weakly supervised training labels. On a large collection of millions of randomly collected English tweets containing emojis, they try to predict the specific emojis from features obtained from the remaining tweet text. We will follow this idea of transfer learning to evaluate it for offensive language detection in German Twitter data together with other transfer learning strategies.

\section{Data and Tasks}
\label{sec:data}

\subsection{GermEval 2018 Shared Task}
Organizers of GermEval 2018 provide training and test datasets for two tasks. 
\textit{Task 1} is a binary classification for deciding whether or not a German tweet contains offensive language (the respective category labels are `offense' and `other'). \textit{Task~2} is a multi-class classification with more fine-grained labels sub-categorizing the same tweets into either `insult', `profanity', `abuse', or `other'.

The training data contains 5,008 manually labeled tweets sampled from Twitter from selected accounts that are suspected to contain a high share of offensive language. Manual inspection reveals a high share of political tweets among those labeled as offensive. These tweets range from offending single Twitter users, politicians and parties to degradation of whole social groups such as Muslims, migrants or refugees.
The test data contains 3,532 tweets. To create a realistic scenario of truly unseen test data, training and test set are sampled from disjoint user accounts.
No standard validation set is provided for the task. To optimize hyper-parameters of our classification models and allow for early stopping to prevent the neural models from overfitting, we created our own validation set. For this, we used the last 808 examples from the provided training set. The remaining first 4,200 examples were used to train our models.

\subsection{Background Knowledge}
Since the provided dataset for offensive language detection is rather small, we investigate the potential of transfer learning to increase classification performance. For this, we use the following labeled as well as unlabeled datasets.

\paragraph{One Million Posts:} A recently published resource of German language social media data has been published by \newcite{Schabus2017}. Among other things, the dataset contains 11,773 labeled user comments posted to the Austrian newspaper website `Der Standard'.\footnote{\url{http://derstandard.at}} Comments have not been annotated for offensive language, but for categories such as \textit{positive/negative sentiment}, \textit{off-topic}, \textit{inappropriate} or \textit{discriminating}.

\paragraph{Twitter:} As a second resource, we use a background corpus of German tweets that were collected using the Twitter streaming API from 2011 to 2017. Since the API provides a random fraction of all tweets (1\%), language identification is performed using `langid.py' \cite{liu2012} to filter for German tweets. For all years combined, we obtain about 18 million unlabeled German tweets from the stream, which can be used as a large, in-domain background corpus.

\section{Text Classification}
\label{sec:model}

In the following section, we describe one linear classification model in combination with specifically engineered features, which we use as a baseline for the classification task. We further introduce a neural network model as a basis for our approach to transfer learning. This model achieves the highest performance for offensive language detection, as compared to our baseline.

\subsection{SVM baseline:} 

\paragraph{\textbf{Model:}} The baseline classifier uses a linear Support Vector Machine \cite{liblinear}, which is suited for a high number of  features. We use a text classification framework for German \cite{Ruppert2017} that has been used successfully for sentiment analysis before.
\paragraph{\textbf{Features:}} We induce token features based on the Twitter background corpus.
Because tweets are usually very short, they are not an optimal source to obtain good estimates on inverse document frequencies (IDF).
To obtain a better feature weighting, we calculate IDF scores based on the Twitter corpus combined with an in-house product review dataset (cf.~ibid.).
From this combined corpus, we compute the IDF scores and 300-dimensional word embeddings \cite{Mikolov2013} for all contained features. 
Following \newcite{Ruppert2017}, we use the IDF scores to obtain the highest-weighted terms per category in the training data. Here, we obtain words like \emph{Staatsfunk, Vasall} (state media, vassal) or \emph{deutschlandfeindlichen} (Germany-opposing) for the category `abuse' and curse words for `insult'. Further, IDF scores are used to weight the word vectors of all terms in a tweet.
Additionally, we employ a polarity lexicon and perform lexical expansion on it to obtain new entries from our in-domain background corpus that are weighted on a `positive--negative' continuum. Lexical expansion is based on distributional word similarity as described in \newcite{Kumar.2016}. 

\subsection{BiLSTM-CNN for Text Classification}

\begin{figure}[h]
\centering
\includegraphics[width=0.5\textwidth,trim={0.6cm 0 0 1cm},clip]{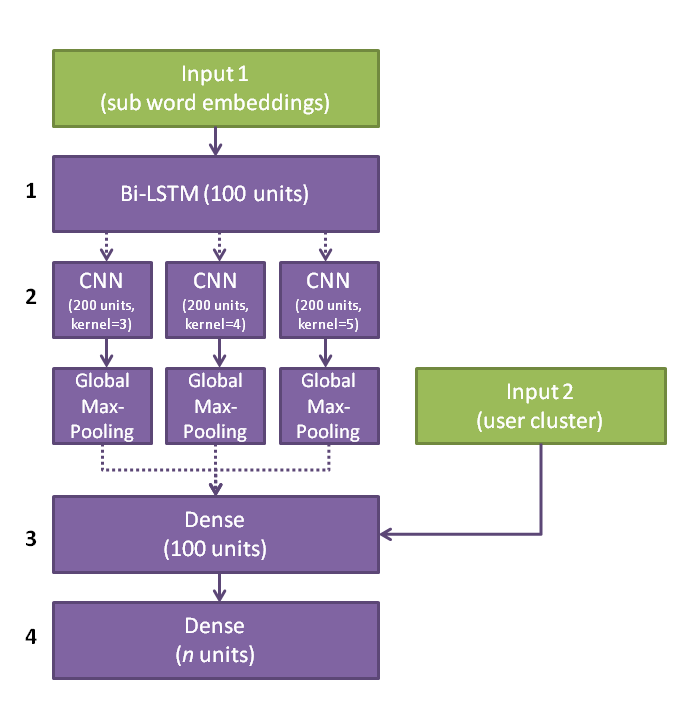}
\caption{BiLSTM-CNN model architecture. We use a combination of recurrent and convolutional cells for learning. As input, we rely on (sub-)word embeddings. The final architecture also includes clustering information obtained from Twitter user ids. Dotted lines indicate dropout with rate 0.5 between layers. The last dense layer contains $n$ units for prediction of the probability of each of the $n$ classification labels per task.} 
\label{fig:architecture}
\end{figure}

\paragraph{\textbf{Model:}} For transfer learning, we rely on a neural network architecture implemented in the Keras framework for Python.\footnote{\url{https://keras.io}} Our model (see Fig.~\ref{fig:architecture}) combines a bi-directional LSTM layer \cite{Hochreiter.1997} with 100 units followed by three parallel convolutional layers (CNN), each with a different kernel size $k\in {3,4,5}$, and a filter size 200. The outputs of the three CNN blocks are max-pooled globally and concatenated. Finally, features encoded by the CNN blocks are fed into a dense layer with 100 units, followed by the prediction layer. Except for this final layer which uses Softmax activation, we rely on LeakyReLU activation \cite{Maas.2013} for the other model layers. For regularization, dropout is applied to the LSTM layer and to each CNN block after global max-pooling (dropout rate 0.5). For training, we use the Nesterov Adam optimization and categorical cross-entropy loss with a learning rate of 0.002.
The intuition behind this architecture is that the recurrent LSTM layer can serve as a feature encoder for general language characteristics from sequences of semantic word embeddings. The convolutional layers on top of this can then encode category related features delivered by the LSTM while the last dense layers finally fine-tune highly category-specific features for the actual classification task.
\paragraph{\textbf{Features:}}
As input, we feed 300-dimensional word embeddings obtained from fastText \cite{Bojanowski.2017} into our model. Since fastText also makes use of sub-word information (character n\hbox{-}grams), it has the great advantage that it can provide semantic embeddings also for words that have not been seen during training the embedding model. We use a model pre-trained with German language data from Wikipedia and Common Crawl provided by \newcite{mikolov2018advances}. 
First, we unify all Twitter-typical user mentions (`@username') and URLs into a single string representation and reduce all characters to lower case. 
Then, we split tweets into tokens at boundaries of changing character classes. As an exception, sequences of emoji characters are split into single character tokens.
Finally, for each token, an embedding vector is obtained from the fastText model.

For offensive language detection in Twitter, users addressed in tweets might be an additional relevant signal. We assume it is more likely that politicians or news agencies are addressees of offensive language than, for instance, musicians or athletes.
To make use of such information, we obtain a clustering of user ids from our Twitter background corpus. From all tweets in our stream from 2016 or 2017, we extract those tweets that have at least two @-mentions and all of the @-mentions have been seen at least five times in the background corpus. Based on the resulting 1.8 million lists of about 169,000 distinct user ids, we compute a topic model with $K=50$ topics using Latent Dirichlet Allocation \cite{Blei.2003}. For each of the user ids, we extract the most probable topic from the inferred user id-topic distribution as cluster id. This results in a thematic cluster id for most of the user ids in our background corpus grouping together accounts such as American or German political actors, musicians, media websites or sports clubs (see Table~\ref{tab:users}). 
For our final classification approach, cluster ids for users mentioned in tweets are fed as a second input in addition to (sub-)word embeddings to the penultimate dense layer of the neural network model.

\begin{table}[]
\centering
\caption{Examples of Twitter user clusters}
\label{tab:users}
\begin{tabular}{@{}lp{6cm}@{}}
\toprule
\textbf{Cluster} & \textbf{Accounts}                                                       \\ \midrule
26               & breitbartnews, realdonaldtrump, jrcheneyjohn, lindasuhler, barbmuenchen \\
28               & dagibee, lilyachty, youngthug, chrisbrown, richthekid                   \\
40               & bvb, fcbayern, dfb, young, team                                         \\
44               & spdde, cdu, gruenen, martinschulz, fdp, dielinke
\\
50               & tagesschau, spiegelonline, zdf, zeitonline, janboehm                    \\ \bottomrule
\end{tabular}
\end{table}

\section{Transfer Learning}
\label{sec:transfer}

As mentioned earlier, we investigate potential strategies for transfer learning to achieve optimal performance. For this, we compare three different methods to pre-train our model with background data sets. We also compare three different strategies to combat `catastrophic forgetting' during training on the actual target data. 

\subsection{Background Knowledge}

For a transfer learning setup, we need to specify a task to train the model and prepare the corresponding dataset. We compare the following three methods.

\paragraph{Supervised near-category transfer:} As introduced above, the `One Million Post' corpus provides annotation labels for more than 11,000 user comments. Although there is no directly comparable category capturing `offensive language' as defined in the shared task, there are two closely related categories. From the resource, we extract all those comments in which a majority of the annotators agree that they contain either `inappropriate' or `discriminating' content, or none of the aforementioned. We treat the first two cases as examples of `offense' and the latter case as examples of `other'. This results in 3,599 training examples (519 offense, 3080 other) from on the `One Million Post' corpus. We conduct pre-training of the neural model as a binary classification task (similar to the Task~1 of Germ\-Eval~2018)

\paragraph{Weakly-supervised emoji transfer:} Following the approach of \newcite{Felbo.2017}, we constructed a weakly-supervised training dataset from our Twitter background corpus. From all tweets posted between 2013 and 2017, we extract those containing at least one emoji character. In the case of several emojis in one tweet, we duplicate the tweet for each unique emoji type. Emojis are then removed from the actual tweets and treated as a label to predict by the neural model. This results in a multi-class classification task to predict the right emoji out of 1,297 different ones. Our training dataset contains 1,904,330 training examples.

\paragraph{Unsupervised topic transfer:} As a final method, we create a training data set for transfer learning in a completely unsupervised manner. For this, we compute an LDA clustering with $K=1,000$ topics\footnote{For LDA, we used Mallet (\url{http://mallet.cs.umass.edu}) with Gibbs Sampling for 1,000 iterations and priors  $\alpha=10 / K$ and $\beta=0.01$.} on 10 million tweets sampled from 2016 and 2017 from our Twitter background corpus containing at least two meaningful words (i.e. alphanumeric sequences that are not stopwords, URLs or user mentions). Tweets also have been deduplicated before sampling. From the topic-document distribution of the resulting LDA model, we determined the majority topic id for each tweet as a target label for prediction during pre-training our neural model. Pre-training of the neural model was conducted on the 10 million tweets with batch size 128 for 10 epochs.

\subsection{Transfer Learning Strategies}

Once the neural model has been pre-trained on the above-specified targets and corresponding datasets, we can apply it for learning our actual target task. For this, we need to remove the final prediction layer of the pre-trained model (i.e. Layer~4 in Fig.~\ref{fig:architecture}), and add a new dense layer for prediction of one of the actual label sets (two for Task~1, four for Task~2).
The training for the actual Germ\-Eval tasks is conducted with batch size 32 for up to 50 epochs.
To prevent the aforementioned effect of forgetting pre-trained knowledge during this task-specific model training, we evaluate three different strategies.

\paragraph{Gradual unfreezing (GU):} In \newcite{Howard.2018}, gradual unfreezing of pre-trained model weights is proposed as one strategy to mitigate forgetting. The basic idea is to initially freeze all pre-trained weights of the neural model and keep only the newly added last layer trainable (i.e. Layer 4 in Fig.~\ref{fig:architecture}). After training that last layer for one epoch on the GermEval training data, the next lower frozen layer is unfrozen and training will be repeated for another epoch. This will be iterated until all layers (4 to 1) are unfrozen.

\paragraph{Single bottom-up unfreezing (BU):} Following the approach of \newcite{Felbo.2017}, we do not iteratively unfreeze all layers of the model, but only one at a time. First, the newly added final prediction layer is trained while all other model weights remain frozen. Training is conducted for up to 50 epochs. The best performing model during these epochs with respect to our validation set is then used in the next step of fine-tuning the pre-trained model layers. For the bottom-up strategy, we unfreeze the lowest layer (1) containing the most general knowledge first, then we continue optimization with the more specific layers (2 and 3) one after the other. During fine-tuning of each single layer, all other layers remain frozen and training is performed for 50 epochs selecting the best performing model at the end of each layer optimization. In a final round of fine-tuning, all layers are unfrozen.

\paragraph{Single top-down unfreezing (TU):} This proceeding is similar the one described above, but inverts the order of unfreezing single layers from top to bottom sequentially fine-tuning layers 4, 3, 2, 1 individually, and all together in a final round.

\paragraph{Baseline (Pre-train only):} All strategies are compared to the baseline of no freezing of model weights, but training all layers at once directly after pre-training with one of the three transfer datasets.

\begin{table}
\caption{Transfer learning performance (Task 1)}
\label{tab:transfer1}
\resizebox{0.48\textwidth}{!}{
\begin{tabular}{@{}llrr@{}}
\toprule
\textbf{Transfer}            & \textbf{Strategy} & \textbf{F1}    & \textbf{Accuracy}   \\ \midrule
None                           & -                 & 0.709          & 0.795          \\ \midrule
\multirow{4}{*}{Category}        & Pre-train only     & \textbf{0.712} & \textbf{0.809} \\
                            & GU                & 0.702          & 0.796          \\
                            & BU                & 0.709          & 0.802          \\
                            & TU                & 0.711          & 0.799          \\ \midrule
\multirow{4}{*}{Emoji}      & Pre-train only     & 0.720          & 0.811          \\
                            & GU                & 0.708          & 0.807          \\
                            & BU                & \textbf{0.739} & \textbf{0.817} \\
                            & TU                & 0.725          & 0.814          \\ \midrule
\multirow{4}{*}{Topic} & Pre-train only     & 0.733          & 0.817          \\
                            & GU                & 0.712          & 0.801          \\
                            & BU                & \textbf{0.753} & \textbf{0.828} \\
                            & TU                & 0.732          & 0.817          \\ \bottomrule
\end{tabular}}
\end{table}

\section{Evaluation}
\label{sec:eval}

Since there is no prior state-of-the-art for the Germ\-Eval Shared Task 2018 dataset, we evaluate the performance of our neural model compared to the baseline SVM architecture.
We further compare the different tasks and strategies for transfer learning introduced above and provide some first insights on error analysis.

\paragraph{Transfer learning:} First, we evaluate the performance of different transfer learning datasets and strategies. Tables~\ref{tab:transfer1} and \ref{tab:transfer2} show that we achieve best performances for both tasks on our validation set by pre-training our neural model on the large Twitter datasets.\footnote{For the binary classification Task~1, we report precision (P), recall (R), and F1 for the targeted positive class `offense'. During training, we also optimized for binary F1. For the multi-class classification Task~2, we report macro-F1 (average of precision, recall, and F1 of all individual four categories). During training, we also optimized for macro-F1. All reported results are average values obtained from 10 repeated runs of model training.}
The two approaches, emoji and topic transfer, substantially improve the classification performance compared to not using transfer learning at all (`None'). In contrast, pre-training on the annotated dataset from the `One Million Posts' corpus does only lead to minor improvements. Comparing the three different strategies to reduce negative effects of forgetting in transfer learning, the strategy of unfreezing single layers during training from the lowest layers to the top of the model architecture (BU) performs best, especially in conjunction with the pre-training on the large Twitter datasets. 
For these setups, the model can take full advantage of learning language regularities from generic to more task-specific features in its different layers.
The other strategies (GU, TU) do not perform better than pre-training the neural model and then immediately training the entire network on the actual task (`Pre-train only').

\begin{table}
\centering
\caption{Transfer learning performance (Task 2)}
\label{tab:transfer2}
\resizebox{0.48\textwidth}{!}{%
\begin{tabular}{@{}llrr@{}}
\toprule
\textbf{Transfer}            & \textbf{Strategy} & \textbf{F1}    & \textbf{Accuracy}   \\ \midrule
None                           & -                 & 0.578          & 0.747          \\ \midrule
\multirow{4}{*}{Category}        & Pre-train only     & 0.578          & 0.755          \\
                            & GU                & 0.560          & 0.751          \\
                            & BU                & 0.580          & 0.750          \\
                            & TU                & \textbf{0.581} & \textbf{0.759} \\ \midrule
\multirow{4}{*}{Emoji}      & Pre-train only     & 0.572          & 0.756          \\
                            & GU                & 0.564          & 0.756          \\
                            & BU                & 0.577          & \textbf{0.764} \\
                            & TU                & \textbf{0.592} & 0.757          \\ \midrule
\multirow{4}{*}{Topic} & Pre-train only     & 0.597          & 0.762          \\
                            & GU                & 0.590          & 0.755          \\
                            & BU                & \textbf{0.607} & \textbf{0.764} \\
                            & TU                & 0.582          & 0.764          \\ \bottomrule 
\end{tabular}%
}
\end{table}

\paragraph{Final results:} 
Tables~\ref{tab:finalresults} and \ref{tab:finalresults2} show the final results for the two offensive language detection tasks on the official test set.
We compare the baseline SVM model
with the BiLSTM-CNN neural model with the best performing transfer learning setup (BU). Additionally, we show the results when adding cluster information from users addressed in tweets (cf. Section~\ref{sec:model}). Due to the fact that training and validation data were sampled from a different user account population than the test dataset (cf. Section~\ref{sec:data}), evaluation scores on the official test data are drastically lower than scores achieved on our validation set during model selection.

Compared to the already highly tweaked SVM baseline, our BiLSTM-CNN model architecture with topic transfer delivers comparable results for identifying offensive language in Task 1 and significantly improved results for Task 2. The SVM achieves a high precision but fails to identify many offensive tweets, which especially in Task~2 negatively affects the recall. 

In contrast, topic transfer leads to a significant improvement, especially for Task 2. Performance gains mainly stem from increased recall due to the background knowledge incorporated into the model.
We assume that not only language regularities are learned through pre-training but that also some aspects relevant for offensive language already are grouped together by the LDA clusters used for pre-training.

As a second task-specific extension of our text classification, we feed cluster information for users addressed in tweets into the process.
Here the results are mixed. 
While this information did not lead to major performance increases on our validation set (not shown), the improvements for the official test set are quite significant.
For Task 1, the performance score increases several percentage points up to 75.2\% F1 (Accuracy 77.5\%). For Task~2, increases are still quite remarkable, although the absolute performance of this multi-class problem with 52.7\% F1 (Accuracy 73.7\%) is rather moderate.
From these results, we infer that thematic user clusters apparently contribute a lot of information to generalize an offensive language detection model to unseen test data.

\begin{table*}
\centering
\caption{Offensive language detection performance \% (Task 1)}
\label{tab:finalresults}
\resizebox{\textwidth}{!}{%
\begin{tabular}{@{}llllllllllll@{}}
\toprule
\textbf{Model}        & \textbf{RunID} & \multicolumn{3}{l}{\textbf{Offense}}                               & \multicolumn{3}{l}{\textbf{Other}}                                 & \multicolumn{4}{l}{\textbf{Average (official rank score)}}                         \\ 
                      &                & P                    & R                    & F1                    & P                    & R                    & F1                    & P                    & R                    & F1                    & Acc. \\ \midrule
Baseline SVM          & coarse\_1      & \textbf{71.52}                & 46.17                & 56.12                & 76.52                & \textbf{90.52}                & \textbf{82.93}                & 74.02                & 68.34                & 71.07                & 75.42    \\
BiLSTM-CNN            &                & \multicolumn{1}{l}{} & \multicolumn{1}{l}{} & \multicolumn{1}{l}{} & \multicolumn{1}{l}{} & \multicolumn{1}{l}{} & \multicolumn{1}{l}{} & \multicolumn{1}{l}{} & \multicolumn{1}{l}{} & \multicolumn{1}{l}{} &          \\
+ Topic transfer & coarse\_2      & 66.30                & 49.75                & 56.84                & 77.03                & 86.95                & 81.69                & 71.67                & 68.35                & 69.97                & 74.29    \\
+ User-cluster        & coarse\_3      & 66.29                & \textbf{68.89}                & \textbf{67.56}                & \textbf{83.62}                & 81.93                & 82.77                & \textbf{74.96}                & \textbf{75.41}                & \textbf{75.18}                & \textbf{77.49}    \\ \bottomrule
\end{tabular}%
}
\end{table*}

\begin{table*}
\centering
\caption{Offensive language detection performance \% (Task 2)}
\label{tab:finalresults2}
\resizebox{\textwidth}{!}{%
\begin{tabular}{@{}llllllllll@{}}
\toprule
\textbf{Model}   & \textbf{RunID} & \textbf{Abuse} & \textbf{Insult} & \textbf{Other} & \textbf{Profanity} & \multicolumn{4}{l}{\textbf{Average (official rank score)}}        \\ \midrule
                 &                & F              & F               & F              & F                  & P              & R              & F              & Acc.           \\
Baseline SVM     & fine\_1~~~~~~~~        & 46.10          & 21.12           & 82.88          & 3.92               & 50.92          & 37.27          & 43.04          & 70.44          \\
BiLSTM-CNN       &                &                &                 &                &                    &                &                &                &                \\
+ Topic transfer~~~~~~~~ & fine\_2        & 51.96          & \textbf{40.18}  & 84.26          & 15.58              & 51.06          & 46.07          & 48.44          & 72.79          \\
+ User cluster   & fine\_3        & \textbf{53.25} & 39.46           & \textbf{84.85} & \textbf{29.63}     & \textbf{56.85} & \textbf{49.13} & \textbf{52.71} & \textbf{73.67} \\ \bottomrule
\end{tabular}%
}
\end{table*}

\paragraph{Error analysis:}
Accuracy values for German offensive language detection around 75\% signal some room for improvement in future work. What are the hard cases for classifying offensive language? We look at false positive (FP) and false negatives (FN) for Task~1. In our validation set, the ratio of FP and FN is about 60:40, which means our classifier slightly more often assumes offensive language than there is actually in the data compared to cases in which it misses to recognize offensive tweets. Looking more qualitatively into FP examples, we can see a lot of cases which actually express a very critical opinion and/or use harsh language, but are not unequivocal insults. Another group of FP tweets does not express insults directly but formulates offensive content as a question. In other cases, it is really dependent on context whether a tweet addressing a specific group uses that group signifier actually with a derogatory intention (e.g. calling people \textit{`Jew'}, \textit{`Muslim'}, or \textit{`Communist'}).
For FN tweets, we can identify insults that are rather subtle. They do not use derogatory vocabulary but express loathing by dehumanizing syntax (e.g. \textit{`das was uns regiert'} where the definite gender-neutral article `das' refers to the German chancellor), metaphor (\textit{`Der ist nicht die hellste Kerze'}, i.e. `he is not the brightest light') or insinuating an incestuous relationship of some persons parents (\textit{`Hier dr\"angt sich der Verdacht auf, das die Eltern der beiden Geschwister waren'}). Another repeatedly occurring FN case are tweets expressing suspicion against the government, democratic institutions, the media or elections. While those tweets certainly in most cases origin from a radical right-wing worldview and can be considered as abusive against democratic values, their language is not necessarily offensive per se.
This more qualitative look into the data opens up some directions to improve offensive language detection incorporating technologies that are able to capture such more subtle insults as well as handling cases of questions and harsh but still not insulting critique.

\section{Conclusion}
\label{sec:discussion}

In this paper, we presented our neural network text classification approach for offensive language detection on the GermEval 2018 Shared Task dataset. We used a combination of BiLSTM and CNN architectures for learning. As task-specific adaptations of standard text classification, we evaluated different datasets and strategies for transfer learning, as well as additional features obtained from users addressed in tweets. The coarse-grained offensive language detection could be realized to a much better extent than the fine-grained task of separating four different categories of insults (accuracy 77.5\% vs. 73.7\%). 
From our experiments, four main messages can be drawn: 
\begin{enumerate}
    \item Transfer learning of neural networks architectures can improve offensive language detection drastically.
    \item Transfer learning should be conducted on as much data as possible regarding availability and computational resources. We obtained best results in a completely unsupervised and task-agnostic pre-training setup on in-domain data. During pre-training, we predicted the primary topics of tweets obtained by an LDA process, which previously clustered our background dataset of 10 million tweets into 1,000 topics.
    \item To mitigate the effect of `catastrophic forgetting' in transfer learning, it is advised to train and optimize the different layers of the neural network model separately. 
    In our experiments on models pre-trained on large Twitter datasets, the bottom-up approach of training from the lowest to the top layer performed significantly better than all other tested strategies to freeze model weights during learning.
    \item User mentions in tweets can contribute a lot of information to the classifier since some accounts are much more likely to be targeted by offensive language than others. Clustering users thematically allows including information from users not seen during training.
\end{enumerate}
The fact that our unsupervised, task-agnostic pre-training by LDA topic transfer performed best suggests that this approach will also contribute beneficially to other text classification tasks such as sentiment analysis.
Thus, in future work, we plan to evaluate our approach with regard to such other tasks. 
We also plan to evaluate more task-agnostic approaches for transfer learning, for instance employing language modeling as a pre-training task. 

\newpage
\paragraph{Acknowledgements:} The paper was supported by BWFG Hamburg within the ``Forum 4.0'' project as part of the \textit{ahoi.digital} funding line, and by DAAD via a WISE stipend. 

\bibliographystyle{konvens2018}
\bibliography{references}

\begin{thebibliography}{}

\bibitem[\protect\citename{Blei \bgroup et al.\egroup }2003]{Blei.2003}
David~M. Blei, Andrew~Y. Ng, and Michael~I. Jordan.
\newblock 2003.
\newblock Latent dirichlet allocation.
\newblock {\em Journal of Machine Learning Research}, 3:993--1022.

\bibitem[\protect\citename{Bojanowski \bgroup et al.\egroup
  }2017]{Bojanowski.2017}
Piotr Bojanowski, Edouard Grave, Armand Joulin, and Tomas Mikolov.
\newblock 2017.
\newblock Enriching word vectors with subword information.
\newblock {\em Transactions of the Association for Computational Linguistics},
  5:135--146.

\bibitem[\protect\citename{Fan \bgroup et al.\egroup }2008]{liblinear}
Rong-En Fan, Kai-Wei Chang, Cho-Jui Hsieh, Xiang-Rui Wang, and Chih-Jen Lin.
\newblock 2008.
\newblock {LIBLINEAR}: A library for large linear classification.
\newblock {\em Journal of Machine Learning Research}, 9:1871--1874.

\bibitem[\protect\citename{Felbo \bgroup et al.\egroup }2017]{Felbo.2017}
Bjarke Felbo, Alan Mislove, Anders S{\o}gaard, Iyad Rahwan, and Sune Lehmann.
\newblock 2017.
\newblock Using millions of emoji occurrences to learn any-domain
  representations for detecting sentiment, emotion and sarcasm.
\newblock In {\em Proceedings of the 2017 Conference on Empirical Methods in
  Natural Language Processing}, pages 1615--1625, Copenhagen, Denmark. ACL.

\bibitem[\protect\citename{Hochreiter and Schmidhuber}1997]{Hochreiter.1997}
Sepp Hochreiter and J\"{u}rgen Schmidhuber.
\newblock 1997.
\newblock Long short-term memory.
\newblock {\em Neural Computation}, 9(8):1735--1780.

\bibitem[\protect\citename{Howard and Ruder}2018]{Howard.2018}
Jeremy Howard and Sebastian Ruder.
\newblock 2018.
\newblock Universal language model fine-tuning for text classification.
\newblock In {\em Proceedings of the 56th Annual Meeting of the Association for
  Computational Linguistics (Volume 1: Long Papers)}, pages 328--339,
  Melbourne, Australia. ACL.

\bibitem[\protect\citename{Kim}2014]{Kim.2014}
Yoon Kim.
\newblock 2014.
\newblock Convolutional neural networks for sentence classification.
\newblock In {\em Proceedings of the 2014 Conference on Empirical Methods in
  Natural Language Processing}, pages 1746--1751, Doha, Qatar. ACL.

\bibitem[\protect\citename{Krei{\ss}el \bgroup et al.\egroup
  }2018]{InstituteforStrategicDialogue.2018}
Philip Krei{\ss}el, Julia Ebner, Alexander Urban, and Jakob Guhl.
\newblock 2018.
\newblock {\em {Hass auf Knopfdruck: Rechtsextreme Trollfabriken und das
  {\"O}kosystem koordinierter Hasskampagnen im Netz}}.
\newblock Institute for Strategic Dialogue, London, UK.

\bibitem[\protect\citename{Kumar \bgroup et al.\egroup }2016]{Kumar.2016}
Ayush Kumar, Sarah Kohail, Amit Kumar, Asif Ekbal, and Chris Biemann.
\newblock 2016.
\newblock {IIT-TUDA at SemEval-2016 Task 5}: Beyond sentiment lexicon:
  Combining domain dependency and distributional semantics features for aspect
  based sentiment analysis.
\newblock In {\em Proceedings of the 10th International Workshop on Semantic
  Evaluation}, pages 1129--1135, San Diego, CA, USA. ACL.

\bibitem[\protect\citename{Lui and Baldwin}2012]{liu2012}
Marco Lui and Timothy Baldwin.
\newblock 2012.
\newblock langid.py: An off-the-shelf language identification tool.
\newblock In {\em Proceedings of the 50th Annual Meeting of the Association for
  Computational Linguistics, Demo Session}, pages 25--30, Jeju, Korea. ACL.

\bibitem[\protect\citename{Maas \bgroup et al.\egroup }2013]{Maas.2013}
Andrew~L. Maas, Awni~Y. Hannun, and Andrew~Y. Ng.
\newblock 2013.
\newblock Rectifier nonlinearities improve neural network acoustic models.
\newblock In {\em ICML Workshop on Deep Learning for Audio, Speech, and
  Language Processing}. Atlanta, GA, USA.

\bibitem[\protect\citename{Mikolov \bgroup et al.\egroup }2013]{Mikolov2013}
Tomas Mikolov, Kai Chen, Greg Corrado, and Jeffrey Dean.
\newblock 2013.
\newblock Efficient estimation of word representations in vector space.
\newblock In {\em Workshop at International Conference on Learning
  Representations (ICLR)}, pages 1310--1318, Scottsdale, AZ, USA.

\bibitem[\protect\citename{Mikolov \bgroup et al.\egroup
  }2018]{mikolov2018advances}
Tomas Mikolov, Edouard Grave, Piotr Bojanowski, Christian Puhrsch, and Armand
  Joulin.
\newblock 2018.
\newblock {Advances in Pre-Training Distributed Word Representations}.
\newblock In {\em Proceedings of the 11th International Conference on Language
  Resources and Evaluation}, Miyazaki, Japan. ELRA.

\bibitem[\protect\citename{Nobata \bgroup et al.\egroup }2016]{Nobata.2016}
Chikashi Nobata, Joel Tetreault, Achint Thomas, Yashar Mehdad, and Yi~Chang.
\newblock 2016.
\newblock Abusive language detection in online user content.
\newblock In {\em Proceedings of the 25th International Conference on World
  Wide Web}, pages 145--153, Montreal, Canada. International World Wide Web
  Conferences Steering Committee.

\bibitem[\protect\citename{Ross \bgroup et al.\egroup }2016]{Ross.2016}
Bj\"orn Ross, Michael Rist, Guillermo Carbonell, Ben Cabrera, Nils Kurowsky,
  and Michael Wojatzki.
\newblock 2016.
\newblock Measuring the reliability of hate speech annotations: The case of the
  {European} refugee crisis.
\newblock In {\em Proceedings of 3rd Workshop on Natural Language Processing
  for Computer-Mediated Communication}, pages 6--9, Bochum, Germany.

\bibitem[\protect\citename{Ruppert \bgroup et al.\egroup }2017]{Ruppert2017}
Eugen Ruppert, Abhishek Kumar, and Chris Biemann.
\newblock 2017.
\newblock {LT-ABSA: An} extensible open-source system for document-level and
  aspect-based sentiment analysis.
\newblock In {\em Proceedings of the GSCL GermEval Shared Task on Aspect-based
  Sentiment in Social Media Customer Feedback}, pages 55--60, Berlin, Germany.

\bibitem[\protect\citename{Schabus \bgroup et al.\egroup }2017]{Schabus2017}
Dietmar Schabus, Marcin Skowron, and Martin Trapp.
\newblock 2017.
\newblock One million posts: A data set of {G}erman online discussions.
\newblock In {\em Proceedings of the 40th International Conference on Research
  and Development in Information Retrieval}, pages 1241--1244, Tokyo, Japan.

\bibitem[\protect\citename{Schmidt and Wiegand}2017]{Schmidt.2017}
Anna Schmidt and Michael Wiegand.
\newblock 2017.
\newblock A survey on hate speech detection using natural language processing.
\newblock In {\em Proceedings of the 5th International Workshop on Natural
  Language Processing for Social Media}, pages 1--10, Valencia, Spain. ACL.

\bibitem[\protect\citename{Spertus}1997]{Spertus.1997}
Ellen Spertus.
\newblock 1997.
\newblock Smokey: Automatic recognition of hostile messages.
\newblock In {\em Proceedings of the 14th National Conference on Artificial
  Intelligence and Ninth Conference on Innovative Applications of Artificial
  Intelligence}, pages 1058--1065, Providence, RI, USA. AAAI Press.

\bibitem[\protect\citename{Xiang \bgroup et al.\egroup }2012]{Xiang.2012}
Guang Xiang, Bin Fan, Ling Wang, Jason Hong, and Carolyn Rose.
\newblock 2012.
\newblock Detecting offensive tweets via topical feature discovery over a large
  scale twitter corpus.
\newblock In {\em Proceedings of the 21st ACM International Conference on
  Information and Knowledge Management}, pages 1980--1984, New York, NY, USA.
  ACM.

\end{thebibliography}

\end{document}